\documentclass[11pt,a4paper]{article}
\usepackage{fullpage}
\usepackage{authblk}
 
\usepackage[utf8]{inputenc} 
\usepackage[T1]{fontenc}    
\usepackage{hyperref}   
\usepackage{url}            
\usepackage{booktabs}      
\usepackage{nicefrac}      
\usepackage{microtype}  

\usepackage{graphicx}
\usepackage{amsthm}
\usepackage{amssymb}
\usepackage{amsmath,bm,dsfont}
\theoremstyle{definition}
\newtheorem{myDef}{Definition}
\usepackage{algorithmic,algorithm}

\title{Deep Metric Learning for Few-Shot Image Classification: \\A Review of Recent Developments}

\author[1,2]{Xiaoxu Li\footnote{Equal contribution}}
\author[3]{Xiaochen Yang$^\ast$}
\author[2]{Zhanyu Ma\footnote{Corresponding author: Zhanyu Ma (email: mazhanyu@bupt.edu.cn)}}
\author[4]{Jing-Hao Xue}
\affil[1]{School of Computer and Communication, Lanzhou University of Technology, China.}
\affil[2]{Pattern Recognition and Intelligent System Laboratory, School of Information and Communication Engineering, Beijing University of Posts and Telecommunications, China.}
\affil[3]{School of Mathematics and Statistics, University of Glasgow, UK.}
\affil[4]{Department of Statistical Science, University College London, UK.}
\date{}

\begin{document}

\maketitle

\begin{abstract}
Few-shot image classification is a challenging problem that aims to achieve the human level of recognition based only on a small number of training images. One main solution to few-shot image classification is deep metric learning. These methods, by classifying unseen samples according to their distances to few seen samples in an embedding space learned by powerful deep neural networks, can avoid overfitting to few training images in few-shot image classification and have achieved the state-of-the-art performance. In this paper, we provide an up-to-date review of deep metric learning methods for few-shot image classification from 2018 to 2022 and categorize them into three groups according to three stages of metric learning, namely learning feature embeddings, learning class representations, and learning distance measures. With this taxonomy, we identify the novelties of different methods and problems they face. We conclude this review with a discussion on current challenges and future trends in few-shot image classification.
\end{abstract}

Image classification is an important task in machine learning and computer vision. With the rapid development of deep learning, recent years have witnessed breakthroughs in this area~\cite{krizhevsky2012ImageNet,simonyan2014very,szegedy2015going,gu2015recent}. Such progress, however, hinges on collecting and labeling a vast amount of data (in the order of millions), which can be difficult and costly. More severely, this learning mechanism is in stark contrast with that of humans, where one or few examples suffice for learning a new concept~\cite{fei2006one}. Therefore, to reduce the data requirement and imitate human intelligence, many researchers started to focus on few-shot classification~\cite{koch2015siamese,vinyals2016matching,santoro2016one}, i.e., learning a classification rule from few (typically 1-5) labeled examples.

The biggest challenge in few-shot classification is a high risk of model overfitting to the few labeled training samples. To alleviate this problem, researchers have proposed various approaches, such as meta-learning methods, transfer learning methods, and metric learning methods. Meta-learning methods train a meta-learner on many different classification tasks to extract generalizable knowledge, which enables rapid learning on a new related task with few examples~\cite{vinyals2016matching,finn2017model}. Transfer learning methods presume shared knowledge between the source and target domains, and fine-tune the model trained on abundant source data to fit few labeled target samples~\cite{rohrbach2013transfer,sun2019meta}. 
Metric learning methods learn feature embeddings~\cite{koch2015siamese} and/or distance measures (or inversely, similarity measures)~\cite{sung2018learning} and classify an unseen sample based on its distance to labeled samples or class representations; samples of the same class are expected to locate close together in the embedding space and samples of different classes should be far apart. Note that the above methods can be applied simultaneously, for example learning feature embeddings of metric learning methods by using a meta-learning strategy~\cite{vinyals2016matching}. 

In this paper, we present a review of recent deep metric learning methods for few-shot image classification. Metric learning methods deserve special attention as they do not require learning additional parameters for new classes once the metric is learned, and thus able to avoid overfitting to the few labeled samples of new classes in few-shot learning. They have also demonstrated impressive classification performance on benchmark datasets. Moreover, in this review we decouple metric learning into three learning stages, namely learning feature embeddings, learning class representations, and learning distance measures. Such decomposition facilitates exchange of ideas between researchers from two underpinning communities: few-shot image classification and deep metric learning. For example, latest developments in learning generalizable feature embeddings can be adopted for few-shot image classification, and the idea of learning prototypes, one type of class representations, can be extended for long-tailed visual recognition~\cite{liu2019large}. 

A number of surveys on few-shot learning~(FSL) have been published or preprinted. \cite{shu2018small}~is the first survey on small sample learning, summarizing methods for different small sample learning scenarios, including zero-shot learning and FSL, and for various tasks, such as image classification, object detection, visual question answering, and neural machine translation. Since the survey was conducted early in 2018, it includes relatively limited work on few-shot classification, particularly metric learning methods. \cite{wang2020generalizing} provides the first comprehensive review on FSL. In addition to defining FSL and distinguishing it from related machine learning problems, the authors discuss FSL from the fundamental perspective of error decomposition in supervised learning and classify all methods in terms of augmenting the training data for reducing the estimation error, learning models from prior knowledge for constraining the hypothesis space and reducing the approximation error, and learning initializations or optimizers which improve the search for the optimal hypothesis within the hypothesis space. The survey has limited coverage on metric learning methods and categorize them all under learning embedding models, which does not fully describe the merits of these methods. \cite{lu2020learning} is another comprehensive survey, reviewing literature over a long period from the 2000s to 2020 as well as summarizing applications of FSL in various fields. It includes early, non-deep approaches of metric learning methods and, since the survey emphasizes on meta-learning methods, categorizes most recent, deep approaches under meta-learning as learning-to-measure. Compared with~\cite{lu2020learning} which links different meta-learning metric learning methods to three classical methods, our review provides a deeper insight into how metric learning methods evolve in order to generalize better and be more applicable in the settings that mimic the reality more closely. Moreover, the rapid development of FSL leads to a considerable amount of methods proposed since the publications of~\cite{wang2020generalizing} and \cite{lu2020learning}. These new approaches have been discussed in this review. \cite{li2021concise} is the latest review on FSL published in 2021, but it is entirely devoted to meta-learning approaches and has very little overlap with our work. In short, this paper provides an up-to-date review of deep metric learning methods for few-shot image classification and a careful examination of different components of these methods to understand their strengths and limitations. 

The rest of this review is organized as follows. Firstly for completeness, in Section~\ref{framework} we give the definition of few-shot classification and introduce the evaluation procedure and commonly used datasets. Secondly, in Section~\ref{survey} we review classical few-shot metric learning algorithms and recent influential works published from 2018 to 2022. In the light of the procedure of metric learning, these methods are classified into learning feature embeddings, learning class representations, and learning distance or similarity measures. Finally, we discuss some remaining challenges in existing methods and directions for further developments in Section~\ref{trends} and conclude this review in Section~\ref{Conclusion}. 

\section{The Framework of Few-Shot Image Classification}\label{framework}

\subsection{Notation and definitions}

We first establish the notation and give a unified definition of various types of few-shot classification by generalizing the definition of few-shot learning \cite{sung2018learning}. 

Few-shot classification involves two datasets, \textbf{base dataset} and \textbf{novel dataset}. The novel dataset is the dataset on which the classification task is performed. The base dataset is an auxiliary dataset used to facilitate the learning of the classifier by transferring knowledge. We use $\mathbb{D}_{base}=\{(X_{i},Y_{i}); X_i \in \mathcal{X}_{base},Y_i \in \mathcal{Y}_{base}\}_{i=1}^{N_{base}}$ to denote the base dataset, where $Y_i$ is the class label of instance $X_i$; in the case of image classification, $X_i$ denotes the feature vector of the $i$th image. The novel dataset is denoted similarly by $\mathbb{D}_{novel}=\{(\tilde{X}_{j},\tilde{Y}_{j});\tilde{X}_j \in \mathcal{X}_{novel},\tilde{Y}_j \in \mathcal{Y}_{novel}\}_{j=1}^{N_{novel}}$. \textbf{$\mathbb{D}_{base}$ and $\mathbb{D}_{novel}$ have no overlap in the label space}, i.e., $\mathcal{Y}_{base} \cap \mathcal{Y}_{novel}=\emptyset$. To train and test the classifier, we split $\mathbb{D}_{novel}$ into the support set $\mathbb{D}_S$ and the query set $\mathbb{D}_Q$. 

\theoremstyle{definition}
\begin{myDef}
Suppose the support set $\mathbb{D}_S$ is available, and the sample size of each class in $\mathbb{D}_S$ is very small (e.g., from 1 to 5). The \textbf{few-shot classification} task aims to learn from $\mathbb{D}_S$ a classifier $f: \mathcal{X}_{novel} \rightarrow \mathcal{Y}_{novel}$ that can correctly classify instances in the query set $\mathbb{D}_Q$. In particular, if $\mathbb{D}_S$ contains $C$ classes and $K$ labeled examples per class, the task is called \textbf{$C$-way $K$-shot classification}; if the sample size of each class in $\mathbb{D}_S$ is one, then the task is called \textbf{one-shot classification}.
\end{myDef} 

Before presenting the next definition, we introduce the concept of domain. A \emph{domain} consists of two components, namely a feature space $\mathcal{X}$ and a marginal distribution $P(X)$ over $\mathcal{X}$~\cite{pan2009survey}.
\begin{myDef}
A few-shot classification task is called \textbf{cross-domain few-shot classification} if the base dataset and the novel dataset come from two different domains, i.e., $\mathcal{X}_{base} \neq \mathcal{X}_{novel}$ or $P(X) \neq P(\tilde{X})$, where $X \in \mathcal{X}_{base}$ and $\tilde{X} \in \mathcal{X}_{novel}$. 
\end{myDef} 

\begin{myDef}
The \textbf{generalized few-shot classification} task aims to learn a classifier $f: \mathcal{X}_{novel} \cup \mathcal{X}_{base} \rightarrow \mathcal{Y}_{novel} \cup \mathcal{Y}_{base}$ that can correctly classify instances in the query set $\mathbb{D}_Q$, where $\mathbb{D}_Q$ includes instance-label pairs from $\mathbb{D}_{base}$ in addition to existing pairs from $\mathbb{D}_{novel}$. 
\end{myDef} 

\subsection{Evaluation procedure of few-shot classification}
We provide a general procedure to evaluate the performance of a classifier for $C$-way $K$-shot classification in Algorithm~\ref{alg:evaluation}. The evaluation procedure includes many episodes (i.e., tasks). In each episode, we first randomly select $C$ classes from the novel label set, and then randomly select $K$ samples from each of the $C$ classes to form a support set and $M$ samples from the remaining samples of those $C$ classes to form a query set. Let $\mathbb{X}^{(e)}$ and $\mathbb{Y}^{(e)}$ denote the set of instances and the set of labels in the query set at the $e$th episode, respectively. A learning algorithm returns a classifier $f(\cdot |\mathbb{D}_{base},\mathbb{D}_S^{(e)})$ upon receiving the base dataset and the $e$th support set, which predicts labels of query instances as $\hat{\mathbb{Y}}^{(e)}=f(\mathbb{X}^{(e)} |\mathbb{D}_{base},\mathbb{D}_S^{(e)})$. Let $a^{(e)}$ denote the classification accuracy on the $e$th episode. The performance of a learning algorithm is measured by the classification accuracy averaged over all episodes. 

\newpage
\begin{algorithm}[t] 
\renewcommand{\algorithmicrequire}{\textbf{Input:}}
\renewcommand\algorithmicensure {\textbf{Steps:}}
\caption{Evaluation procedure of $C$-way $K$-shot classification} 
\begin{algorithmic}[1] 
\REQUIRE ~~\\
$\mathbb{D}_{base}=\{(X_{i},Y_{i}); X_i \in \mathcal{X}_{base},Y_i \in \mathcal{Y}_{base}\}_{i=1}^{N_{base}}$; $\mathbb{D}_{novel}=\{(\tilde{X}_{j},\tilde{Y}_{j});\tilde{X}_j \in \mathcal{X}_{novel},\tilde{Y}_j \in \mathcal{Y}_{novel}\}_{j=1}^{N_{novel}}$; number of episodes $E$.\\
\ENSURE ~~\\
\STATE $\text{e} \leftarrow 0$
\REPEAT
\STATE $\text{e} \leftarrow \text{e}+1$
\STATE  Randomly select $C$ classes from $\mathcal{Y}_{novel}$.
\STATE  Randomly select $K$ samples from each class as the support set $\mathbb{D}_S^{(e)}$.
\STATE Randomly select $M$ samples from the remaining samples of $C$ classes as the query set $\mathbb{D}_S^{(e)}=\{(\mathbb{X}^{(e)},\mathbb{Y}^{(e)})\}$.
\STATE Record predicted labels $\hat{\mathbb{Y}}^{(e)} = f(\mathbb{X}^{(e)} |\mathbb{D}_{base},\mathbb{D}_S^{(e)})$.
\STATE Compute accuracy $a^{(e)}=\frac{1}{M}\sum_{j=1}^M \mathds{1}[\hat{\mathbb{Y}}^{(e)}=\mathbb{Y}^{(e)}]$\footnotemark.
\UNTIL {$\text{e}=E$}
\RETURN{mean accuracy $\frac{1}{E}\sum_{e=1}^{E}a^{(e)}$.}    
\end{algorithmic}
\label{alg:evaluation}
\end{algorithm}
\footnotetext{$\mathds{1}$ denotes the element-wise indicator function.}

\subsection{Datasets for few-shot image classification}

\begin{figure}[t]
\begin{center}
    \includegraphics[trim=1.2cm 0.3cm 0.5cm 0.1cm,clip=TRUE,width=\linewidth]{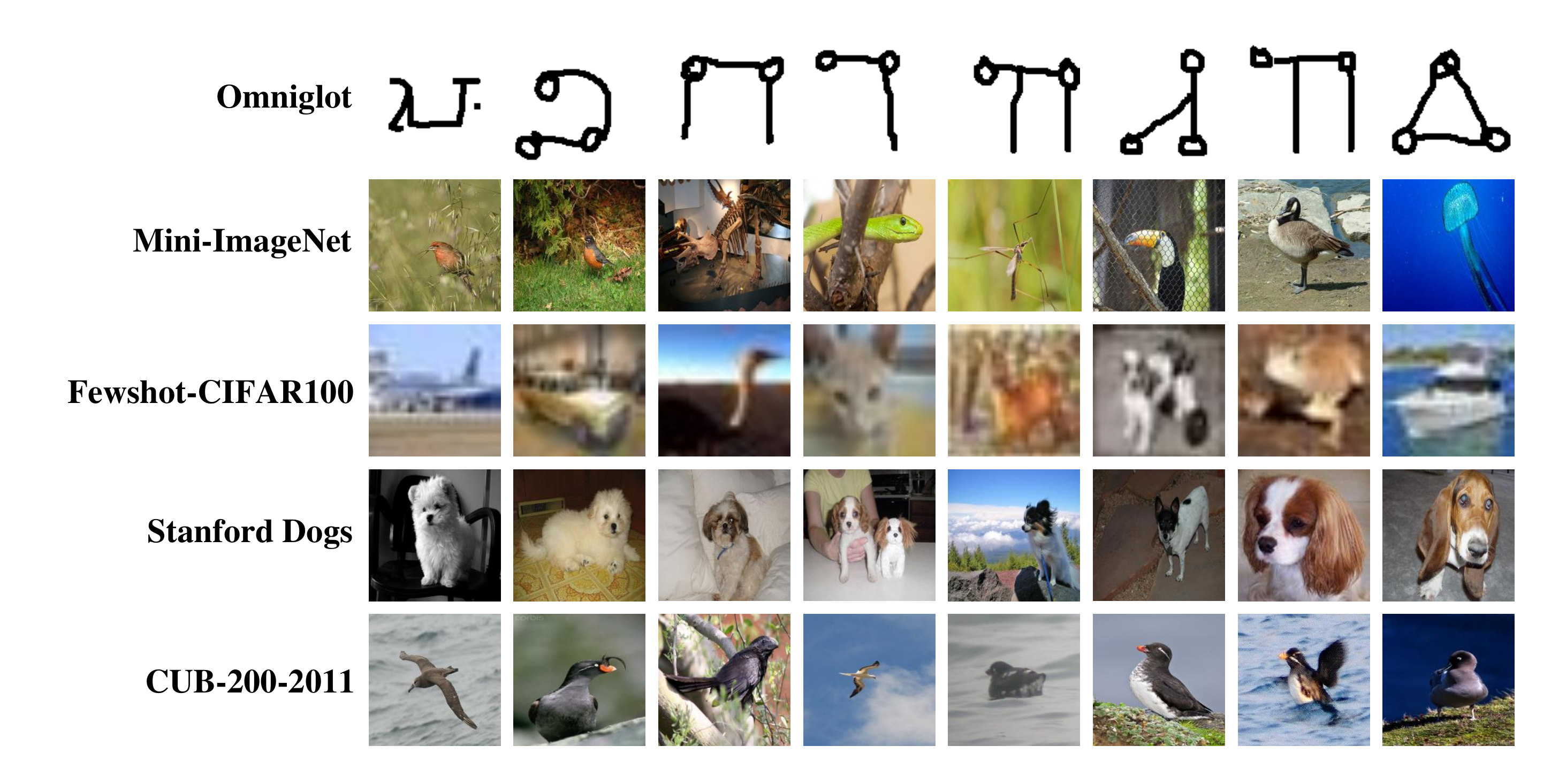}
\end{center}
\caption{Sample images of benchmark datasets for few-shot image classification. Datasets include Onimiglot, Mini-ImageNet, Fewshot-CIFAR100, Stanford Dogs, and CUB-200-2011.}\label{dataset}
\label{fig:example images}
\end{figure}

In this section, we briefly introduce benchmark datasets for few-shot image classification. Statistics of the datasets and commonly used experimental settings are listed below, and sample images are shown in Figure \ref{fig:example images}. 

\begin{itemize}

\item Omniglot~\cite{lake2011one}: one of the most widely used datasets for evaluating few-shot classification algorithms. It contains 1,623 characters from 50 languages. The dataset is often augmented by rotations of 90, 180, 270 degrees, resulting in 6,492 classes, which are split into 4,112 base, 688 validation, and 1,692 novel classes. The validation classes are used for model selection. The dataset is used less often in the latest studies as many methods can attain over 99\% accuracy on the 5-way 1-shot classification task. \end{itemize}

\begin{itemize}
\item Mini-ImageNet and Tiered-ImageNet: another two widely used datasets derived from the ImageNet dataset~\cite{russakovsky2015imagenet}. Mini-ImageNet consists of 100 selected classes with 600 images for each class. This dataset was first proposed by Vinyals et al.~\cite{vinyals2016matching}, but recent studies follow the experimental setting provided by Ravi and Larochelle~\cite{ravi2017optimization}, which splits 100 classes into 64 base, 16 validation, and 20 novel classes. Tiered-ImageNet is a larger dataset with a hierarchical structure~\cite{ren2018meta}. It is constructed from 34 super-classes with 608  classes in total and include 779,165 images. These super-classes are split into 20~base, 6 validation, and 8 novel super-classes, which correspond to 351 base, 97 validation, and 160 novel classes, respectively.

\item CIFAR-FS and FC100: two datasets derived from CIFAR-100~\cite{krizhevsky2009learning}. CIFAR-FS~\cite{bertinetto2019meta} contains 100 classes with 600 images per class, and it is split into 64 base, 16 validation, and 20 novel classes. FC100~\cite{oreshkin2018tadam} divides 100 classes into 20 super-classes, with five classes in each super-class. The dataset is split into 12 base, 4 validation, and 4 novel super-classes. 

\item Stanford Dogs~\cite{khosla2011novel}: one of the benchmark datasets for fine-grained classification task, which contains 120 breeds (classes) of dogs with a total number of 20,580 images. These classes are divided into 70 base, 20 validation, and 30 novel classes. 
\item CUB-200-2011: a fine-grained bird classification dataset, which contains 200 classes and 11,788 images in total. Following \cite{chen2019a}, the dataset is commonly split into 100 base, 50 validation, and 50 novel classes.

\item Mini-ImageNet $ \rightarrow $ CUB: a dataset used for cross-domain few-shot classification.  Mini-ImageNet serves as the base dataset, 50 classes of CUB-200-2011 serve as the validation classes, and the remaining 50 classes serve as novel classes. 

\item Meta-Dataset: a new, large-scale dataset for evaluating few-shot classification methods, particularly cross-domain methods. It initially consists of 10 diverse image datasets~\cite{triantafillou2020meta}, e.g., ImageNet, CUB, and MS COCO~\cite{lin2014microsoft}, and later expanded with three additional datasets~\cite{requeima32fast}. There are two training procedures and two evaluation protocols. In the more commonly used setting of training on all datasets (multi-domain learning)~\cite{requeima32fast,bateni2020improved,li2022cross}, the methods are trained on the official training splits of the first eight datasets, and they are evaluated on the test splits of the same datasets for in-domain performance and the remaining five datasets for out-of-domain performance. The other setting is training only on the Meta-Dataset version of ImageNet (single-domain learning), and evaluating on the test split of ImageNet for in-domain performance and the rest 12 datasets for out-of-domain performance. 
\end{itemize}

\section{Few-Shot Deep Metric Learning Methods}\label{survey}

\begin{figure}[t]
    \centering
    \includegraphics[trim=0cm 0cm 0cm 0.1em,clip=TRUE,width=\textwidth]{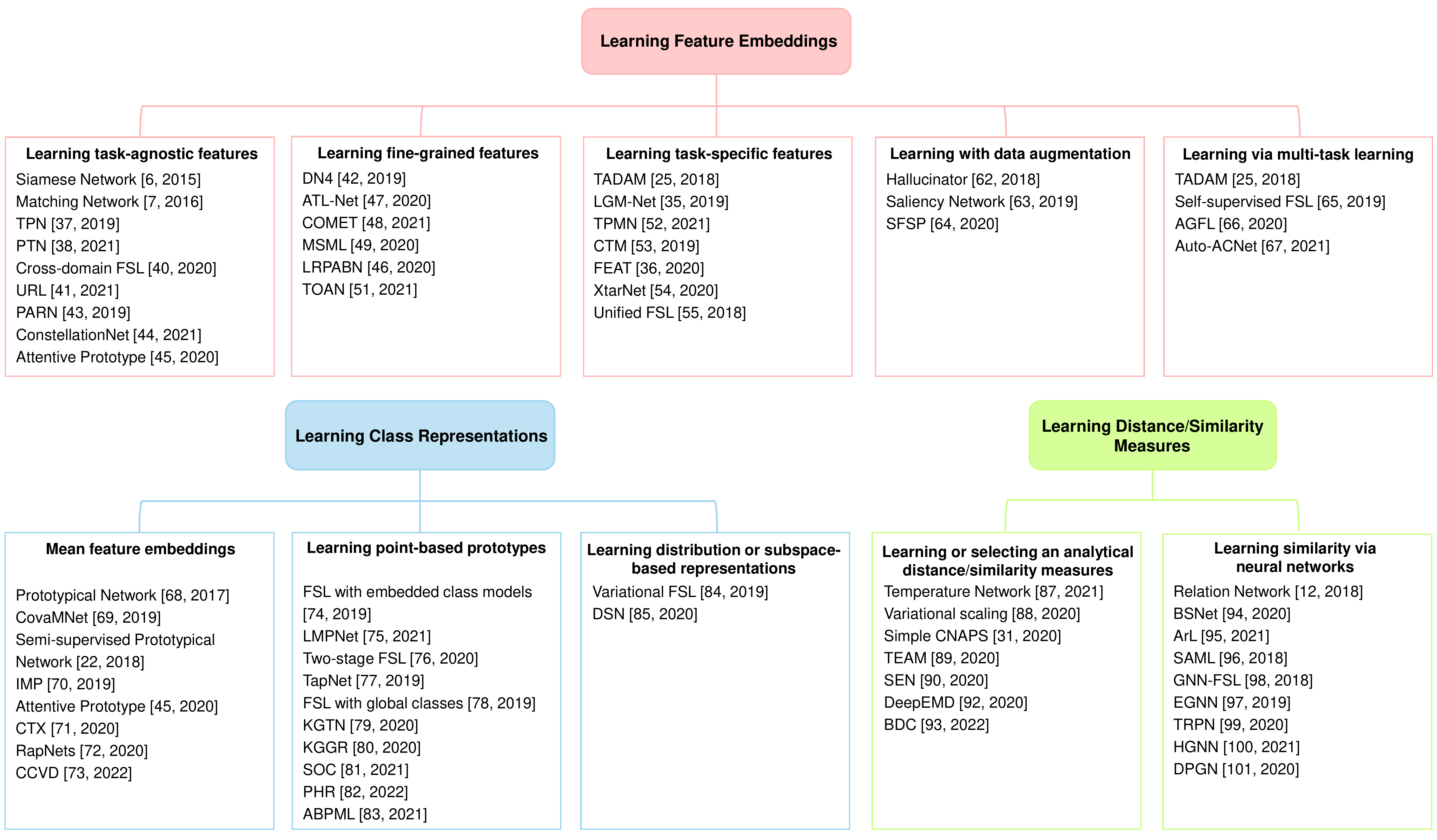}
    \caption{Taxonomy of few-shot deep metric learning methods reviewed in this paper. Some methods contribute to two aspects of metric learning and thus appear twice.}
    \label{fig:taxonomy}
\end{figure}

The goal of supervised metric learning is to learn a distance measure between instance pairs that assigns a small (large, resp.) distance to semantically similar (dissimilar, resp.) instances. In the case of few-shot classification, the metric is learned on the base dataset; query images of the novel class are classified by computing their distances to novel support images with respect to the learned measure, followed by applying a distance-based classifier such as the $k$-nearest neighbor~($k$NN) algorithm. Traditional metric learning methods learn a Mahalanobis distance, which is equivalent to learning a linear transformation of original features~\cite{xing2002distance}. However, in deep metric learning, the distance measure and feature embeddings are often learned separately so as to capture the nonlinear data structure and generate more discriminative feature representations. Moreover, instead of comparing with individual samples, many few-shot metric learning methods compare query samples with class representations such as prototypes and subspaces. In the remainder of this section, we provide a review of representative approaches, which are categorized into three groups according to the aspect they are improving on, namely 1) learning feature embeddings, 2) learning class representations, and 3) learning distance or similarity measures. A summary of these methods is provided in Figure~\ref{fig:taxonomy}. 

\subsection{Learning feature embeddings} \label{sec: feature embedding}

Methods of learning feature embeddings implicitly assume that the network is powerful to extract discriminative features and can generalize well to novel classes. Early approaches aim at a task-agnostic embedding model that is effective for any task. More recently, endeavors are made to learn a task-specific embedding model for better distinguishing the classes at hand. Furthermore, techniques for data augmentation and multi-task learning are leveraged to address the issues of data scarcity and overfitting. 

\subsubsection{Learning task-agnostic features}

The Siamese Convolutional Neural Network~\cite{koch2015siamese} is the first deep metric learning method for one-shot image classification. The Siamese Network, first introduced in~\cite{bromley1993signature}, consists of two sub-networks with identical architectures and shared weights. \cite{koch2015siamese} adopted the VGG-styled convolutional layers as the sub-network to extract high-level features from two images and employed the weighted $L_1$ distance as the distance between the two feature vectors. Weights of the network, as well as those of component-wise distance, are trained using the conventional technique of mini-batch gradient descent.

The Matching Network~\cite{vinyals2016matching} encoded support and query images using different networks in the context of the entire support set, and it first introduced episodic training to few-shot classification. A support image is embedded via a bidirectional LSTM network, which takes account of not only the image itself but also other images in the set; a query image is embedded via an LSTM with an attention mechanism to enable dependency on the support set. However, the sequential nature of bidirectional LSTM results in feature embeddings that will change with different ordering of samples in the support set. This issue can be sidestepped, such as by applying a pooling operation~\cite{li2019lgm} or using self-attention~\cite{ye2020few}. The classification mechanism of Matching Network is suitable for few-shot learning. The network outputs a label distribution by computing a convex combination of one-hot label vectors of all support samples, with coefficients defined by using a softmax over cosine similarities; the class with the highest probability is selected as the predicted class. Another valuable contribution of~\cite{vinyals2016matching} is the episode-based training strategy, which has been adopted by many subsequent works. Following the idea of meta-learning, the training phase on the base dataset should mimic the prediction phase where only few support samples are available. That is, gradient updates should be performed on episodes with $C$ classes randomly sampled from the base label set and $K$ examples for each class. 

The episodic training strategy closes the gap between training and test distributions and thus alleviates the issue of overfitting to few labeled training images. This can also be addressed from a different perspective by utilizing query instances (i.e., excluding query labels) via transductive inference. Transductive Propagation Network (TPN) \cite{liu2019learning} is the first work adopting transductive inference for few-shot learning and introduced the idea of label propagation. Concretely, the network contains a feature embedding module and a graph construction module. The graph construction module, taking feature embeddings as inputs, learns a label propagation graph to exploit the manifold structure of support and query samples. In particular, a $k$NN graph is constructed according to the Gaussian kernel whose length-scale parameters are learned in an example-wise manner. Based on the graph, labels are propagated from the support set to the query set; a closed-form solution of label propagation is used to speed up the prediction procedure. 
The Poisson Transfer Network (PTN)~\cite{huang2020ptn} improved the label propagation scheme by applying the Poisson learning algorithm~\cite{calder2020poisson}, which has been theoretically proved to be more stable and informative, especially in the case of very few labels. PTN is primarily designed for semi-supervised few-shot classification with additional unlabeled data than query samples, and it further enhances the feature embeddings through contrastive self-supervised learning and refines the inference procedure by using a graph-cut method.

The aforementioned methods, designed for classifying novel data from the same domain, degrade when novel data comes from different domains~\cite{chen2019a}. Tseng et al. \cite{tseng2020cross} noticed that this is caused by the large discrepancy between the feature distributions in different domains and proposed to simulate various feature distributions in the training stage as a general solution to enhance the domain generalization ability of metric learning methods. This is achieved by inserting multiple feature-wise transformation layers into the feature extractor; each transformation simulates one distribution, and the hyperparameters of affine transformations can be tuned via a meta-learning approach so that they are optimal to a particular metric learning method and capture the complex variation in feature distributions. Li et al.~\cite{li2021universal} proposed to learn a universal feature representation that works well for multiple domains. The technique of knowledge distillation is applied, where a multi-domain network is learned to generate universal features which align with features from multiple single-domain networks up to a linear transformation.

Motivated by the observation that the interested object may locate only in a region of an image and at different positions across images, a series of improvements on feature embedding have been proposed, such as by learning local features~\cite{wenbin_2019_CVPR} and encoding the position information~\cite{Wu_2019_ICCV}. The former type of methods, while can be applied to generic few-shot image classification, are particularly effective for fine-grained image classification. Thus, we will discuss them separately in the next subsection and only focus on position encoding methods here. Wu et al.~\cite{Wu_2019_ICCV} proposed the Position-Aware Relation Network~(PARN) to reduce the sensitivity of Relation Network (which will be introduced in Section~\ref{sec:learn_CNN_metric}) to the spatial position of semantic objects. PARN adopts deformable convolutional layers to extract more effective features which filter out unrelated information like the background, and a dual correlation attention module to incorporate each spatial position of an image with the global information about the compared image and the image itself, so that the subsequent convolution operations, even subject to local connectivity, can perceive and compare semantic features in different positions. Compared with standard ways of overcoming position sensitivity, such as by using larger kernels or more layers, PARN is more parameter efficient. Xu et al.~\cite{xu2021attentional} proposed the ConstellationNet which extracts part-based features and encodes the spatial relationship between these representations by using self-attention with an explicit, learnable positional encoding. The spatial relationship between different parts of the image has also been encoded in~\cite{wu2020attentive} by using a capsule network.

\subsubsection{Learning task-agnostic features for fine-grained image classification}

Fine-grained image classification aims to distinguish different sub-categories under the same basic-level category. It is particularly challenging due to the subtle differences between different sub-categories and large variance in the same sub-category which may result from variations in the object's pose, scale, rotation, etc. Therefore, for effective classification, several methods have been proposed to extract local features, multi-scale features, and second-order features. 

In deep nearest neighbor neural network~(DN4)~\cite{wenbin_2019_CVPR}, the feature embedding module extracts multiple local descriptors from an image, which are essentially the feature maps learned via CNNs prior to adding the final image-level pooling layer. The classification is performed at an image-to-class level, meaning that the local descriptors from support images of the same class are put into one pool, $k$NNs in each class pool are searched for each query local descriptor, and the total distance over all local descriptors and $k$NNs is the distance between the query image and the corresponding class. The method is shown to be particularly effective on fine-grained datasets, and the idea of learning local descriptors has been adopted in other fine-grained classification methods~\cite{huang2020low}. 
The Adaptive Task-aware Local representations Network~(ATL-Net)~\cite{dong2020learning} improved DN4 by selecting local descriptors with learned thresholds and assigning them different weights based on episodic attention, which brings more flexibility than using $k$NNs and adjusts for the discriminability between classes, respectively. In contrast to learning one feature embedding over spatially local patches, COMET~\cite{cao2021concept} learns multiple embedding functions over various parts of input features. A set of fixed binary masks, termed concepts, are applied to input features to separate an image into human-interpretable segments. For each concept, a feature embedding is learned to map masked features into a new discriminative feature space. The query image is classified according to the distances aggregated from all concept-specific spaces. 

The Multi-Scale Metric Learning~(MSML) network~\cite{jiang2020multi} also constructs multiple feature embeddings, but different from COMET, each embedding corresponds to a different scale of the image. The similarity between support and query features at each scale is computed using the Relation Network.

Huang et al.~\cite{huang2020low} proposed the Low-Rank Pairwise Alignment Bilinear Network~(LRPABN) which aligns features spatially and extracts discriminative, second-order features. After learning first-order features from base images, the method trains a two-layer Multi-Layer Perceptron~(MLP) network with two designed feature alignment losses to transform the positions of image features of a query image to match those of a support image, and designs a low-rank pairwise bilinear pooling layer which adapts the self-bilinear pooling~\cite{wei2019piecewise} to extract second-order features from a pair of support and query images. The classification is performed as in the Relation Network. 
In the follow-up work, \cite{huang2021toan} improves the spatial alignment part by using the cross-channel attention to generate spatially matched support and query features and groups features in the convolutional channel dimension before the pooling layer as each group corresponds to a semantic concept.

\subsubsection{Learning task-specific features}

Methods reviewed in the preceding sections generate the same feature embedding for an image, regardless of the subsequent classification task.  While this avoids the risk of overfitting, these generic features may not be sufficiently discriminative to distinguish novel classes. To this end, task-specific embedding models have been proposed to adapt features to the particular task; it should be noted that the adaptation is learned on the base dataset and does not  involve any re-training on the novel dataset. 

TADAM~\cite{oreshkin2018tadam} is the first metric learning method which explicitly performs task adaptation. Exploiting the technique of conditional batch normalization, it applies a task-specific affine transformation to each convolutional layer of a task-agnostic feature extractor. The task is represented by the mean of class prototypes, and the scale and shift parameters of the affine transformation are generated from a separate network, called the Task Embedding Network (TEN). As TEN introduces more parameters and causes difficulty in optimization, the training scheme is revised to add the standard training, i.e., to distinguish all classes in the base dataset, as an auxiliary task to the episodic training.

Li et al.~\cite{li2019lgm} proposed a meta-learning approach that can adapt weights of Matching Network to novel data. The proposed LGM-Net consists of a meta-learner termed MetaNet and a task-specific learner termed TargetNet. The MetaNet module learns to produce a representation of each task from the support set and construct a mapping from the representation to weights of TargetNet. The TargetNet module, set as the Matching Network, embeds support and query images and performs classification. The proposed meta-learning strategy can be potentially implemented to adapt network parameters of other metric learning methods. Wu et al.~\cite{wu2021task} also proposed to learn task-specific parameters, but they combined the idea with local features. The proposed Task-aware Part Mining Network~(TPMN) learns to generate parameters of filters used for extracting part-based features.

Different from the above two works which generate parameters for task-specific embedding layers, Li et al.~\cite{li2019finding} proposed to modify the generic features output from the task-agnostic embedding layers. A task-specific feature mask is generated from the Category Traversal Module (CTM), which includes a concentrator unit and a projector unit to extract features for intra-class commonality and inter-class uniqueness, respectively. It is noted that CTM can be easily embedded into most few-shot metric learning methods, such as Matching Network, Prototypical Network, and Relation Network; the latter two methods will be introduced in the following sections. Ye et al.~\cite{ye2020few} also proposed to adjust features directly, but instead of applying a mask, set-to-set functions are used to transform a set of task-agnostic features into a set of task-specific ones. These functions can model interactions between images in a set and hence enable co-adaptation of each image. Four set-to-set function approximators are presented in~\cite{ye2020few}, and the one with Transformer, termed FEAT, is shown to be most effective.

Yoon et al.~\cite{yoon2020xtarnet} proposed XtarNet to learn task-specific features for a new setting of generalized few-shot learning, where the model is trained on the base dataset, adapted given the support set of the novel dataset, and used to classify instances from both base and novel classes. XtarNet contains three meta-learners. The MetaCNN module adapts feature embeddings for each task. The MergeNet module produces weights for mixing pre-trained features and meta-learned features. As the classification is performed by comparing the mixed features with class prototypes, the TconNet module adapts prototypes of base and novel classes to improve discriminability. 
Rahman et al.~\cite{rahman2018unified} proposed a unified approach for zero-shot learning, generalized zero-shot learning and few-shot learning, which classifies a query image based on the similarity between its semantic representation and the textual features of each class. The semantic representation is a combination of two parts -- one is a linear combination of base samples' semantic features, and the other one is based on the linear mapping learned from support images.

\subsubsection{Feature learning with data augmentation}

Data augmentation is a strategy that expands the support set in an artificial or model-based way with label preserving transformations, and thus is well-suited when the support samples are limited. One commonly used method is deformation \cite{kulkarni2015deep,ratner2017learning,perez2017effectiveness}, such as cropping, padding, and horizontal flipping. Besides this, generating more training samples \cite{shrivastava2017learning, antoniou2017data} and pseudo labels \cite{ratner2016data} are also popular techniques to augment data. 

In few-shot learning, there is one class of works which places the data augmentation process into a model, that is, they embed a generator that can generate the augmented data to learn or imagine the diversity of data. Wang et al.~\cite{wang2018low} constructed an end-to-end few-shot learning method, in which the training data goes through two streams to output -- one is from the original data to the classifier directly, and the other one is from the original data to a `hallucination' network to augment data and then from the augmented data to classifier. Zhang et al. \cite{zhang2019few} developed a saliency-based data generation strategy. The Saliency Network obtains foregrounds and backgrounds of an image, which are used to achieve the hallucination for the image.  

In~\cite{guan2020zero}, a much simpler feature synthesis strategy is proposed, which synthesizes novel features by perturbing the semantic representations (i.e., word vectors of class labels) and projecting them into the visual feature space. In addition, when learning the projection function, a competitive learning formulation is adopted to push the synthesized sample towards the center of the most likely unseen class and away from that of the second best class.

\subsubsection{Multi-task feature learning}
Besides generating more training data, some works tried to exploit auxiliary information of samples to perform multi-task learning, which creates a regularization effect and helps learn discriminative features. 

As briefly discussed above, TADAM~\cite{oreshkin2018tadam} used an auxiliary task of training a normal global classifier on the base dataset to co-train the few-shot classifier; the task is sampled with a probability during the training process. Gidaris et al. \cite{Gidaris_2019_ICCV} proposed a few-shot learning method combining self-supervised learning. In specific, support samples are artificially rotated to different number of degrees. A shared feature embedding is learned through two branches of networks, one for the original classification task and the other for identifying the rotation degree. 
Zhu et al.~\cite{zhu2020attribute} suggested that base and novel classes, despite being disjoint, can be connected by some visual attributes. Based on this insight, they used attribute learning as an auxiliary task. Visual attributes are provided as additional information during training, and the embedding network is learned to correctly predict both attribute labels and class labels. 
\cite{zhang2021auto} also utilized attribute information, but in a richer way which requires an additional prediction of common and different attributes between an image pair. Moreover, the neural architecture search was first introduced to few-shot learning for automatically identifying the optimal operation from max pooling, convolution, identity mapping, etc for layers in the feature embedding network and attribute learning network.

\subsection{Learning class representations} \label{sec: prototype}
Early few-shot metric learning methods such as Siamese Network and Matching Network classify a query sample by measuring and comparing its distance to support samples. However, since support samples are scarce, they have limited capacity in representing the novel class. To alleviate this issue, some researchers propose to use class prototypes, which serve as reference vectors for each class. Prototypes can be constructed by taking simple or weighted average of feature embeddings, or learned in an end-to-end manner so as to further improve their representation ability. Besides point-based prototypes, some works consider the distribution of each class or use subspaces as class representations. 

\subsubsection{Feature embeddings-based prototypes}

Prototypical Network \cite{snell2017prototypical} is a classical method that performs classification by calculating the Euclidean distance to class prototypes in the learned embedding space. It builds on the hypothesis that there exists an embedding space in which each class can be represented by a single prototype and all instances cluster around the prototype of their corresponding classes. In~\cite{snell2017prototypical}, the prototype of each class is set as the mean of feature embeddings of support samples in the class. Feature embeddings, and thus class prototypes, are learned using episodic training with the objective of minimizing the cross entropy loss. In~\cite{li2019distribution}, the class prototype is represented using the covariance matrix of feature embeddings. A covariance-based metric is also proposed to measure the similarity between the query and the class.

To make use of both labeled support samples and unlabeled samples, Ren et al. \cite{ren2018meta} proposed semi-supervised Prototypical Network, which is the first work of semi-supervised few-shot learning. The method adopts soft $k$-means to compute assignment score of unlabeled samples and computes the prototype as the mean of weighted samples based on assignment scores. 

Considering that the dataset may exhibit multi-modality and multiple prototypes would be more suitable in this scenario, Infinite Mixture Prototypes (IMP) \cite{allen2019infinite} modeled multiple clusters in each class, and each cluster is modeled as a Gaussian distribution. Concretely, the probability that a sample follows the Gaussian distribution of each cluster determines which cluster the sample is assigned into. Moreover, the cluster variance of the Gaussian distributions, which needs to be learned, can affect the number of class prototype and performance of IMP. 

Wu et al.~\cite{wu2020attentive} proposed to compute query-dependent prototypes. An attentive prototype is computed for each query as the weighted average of support samples and the weights are given by the Gaussian kernel with the Euclidean distance between the query and the support samples. As support samples that are more relevant to the query have greater influence on the classification, the method is more robust to outliers in support samples. Query-dependent prototypes have also been studied in CrossTransformers~(CTX)~\cite{doersch2020crosstransformers}, but they are computed separately for each spatial location. In other words, a local region of a query image is compared with an attentive prototype specific to this query and region, and the overall distance between the query and the prototype is the averaged distances over all local regions. Moreover, self-supervised episodes are constructed to train CTX. 

Lu et al.~\cite{lu2020robust} focused on enhancing the robustness of prototypes against outliers and label noises and proposed the Robust attentive profile Networks~(RapNets). The network transforms raw feature embeddings into correlation features in a nonparametric way, and then inputs these features into a parametric bidirectional LSTM and fully-connected network to generate attention scores which serve as weights to combine support images. Moreover, training episodes are revised to include noisy data, and a new evaluation metric is proposed to evaluate the robustness of few-shot classification methods.

Ma et al.~\cite{ma2022fewshot} provided a geometric interpretation of Prototypical Network, regarding it as a Voronoi diagram. Moreover, the authors extended this perspective and proposed the Cluster-to-Cluster Voronoi Diagram~(CCVD), which can ensemble models learned with different data augmentation, built on single or multiple feature transformations, and using linear or nearest neighbor classifier.

\subsubsection{Point-based learnable prototypes}
Ravichaandran et al.~\cite{Ravichandran_2019_ICCV} adopted an implicit way to learn class representation instead of determining class prototypes as in the aforementioned methods. The prototype is modeled as a learnable and parameterized function of feature embedding of labeled samples in the class and is obtained by minimizing a loss which measures the distance between the feature embedding of a sample and the class prototype. Meanwhile, the function is shot free, that is, it allows sample sizes of classes in novel data to be unbalanced. In~\cite{huang2021local}, prototypes are represented as weighted averages of feature embeddings, but different from~\cite{ren2018meta, wu2020attentive} discussed in the previous section, weights are learned end-to-end via episodic training. Moreover, instead of using image-level features, \cite{huang2021local} combines local descriptors of one class following the idea of DN4 and learn multiple weight vectors to generate multiple prototypes per class. Das and Lee proposed a two-stage approach for generating class prototypes~\cite{das2020two}. In the first stage, feature embeddings are learned, from which coarse prototypes of base and novel classes can be obtained from mean embeddings. In the second stage, the novel class prototype is refined through a meta-learnable function of its own prototype and related base prototypes.

Besides the above methods, TapNet \cite{yoon2019tapnet} explicitly modeled class prototypes as learnable parameters. Prototypes and feature embeddings are learned simultaneously on the base dataset following the training procedure of Prototypical Network. In addition, to make prototypes and feature embeddings more specific to the current task, both of them are projected into a new classification space via a linear projection matrix. The projection matrix is obtained by using a linear nulling operation and does not include any learnable parameter. Luo et al. \cite{Luo_2019_ICCV} proposed to learn prototypes of base and novel classes simultaneously by including the support set of novel classes in the training process. In each episode, local prototypes are generated from the sample synthesis module, which aims to increase the diversity of novel classes. They are then used in the registration module to update the global prototypes towards better separability. The query image is classified by searching for the nearest neighbor among global prototypes. As both base and novel prototypes are learned, the method can be readily applied to the generalized few-shot learning setting. 
Chen et al.~\cite{chen2020knowledge} shared the same aim of learning base and novel prototypes, but additionally took advantage of the semantic correlations among these classes. A Knowledge Graph Transfer Network~(KGTN) is proposed, which employs a gated graph neural network to represent class prototypes and correlations as nodes and edges, respectively. By propagating through the graph, information from correlated base classes is used to guide the learning of novel prototypes. This work is extended in \cite{chen2020knowledge-tpami} to the multi-label classification setting, which employs the attention mechanism and an additional graph for learning class-specific feature vectors.
In~\cite{luo2021rectifying}, the Shared Object Concentrator (SOC) algorithm was proposed to learn a series of prototypes for each novel class from local features of support images. The first prototype is learned to have the largest cosine similarity with one of the local features, the second prototype has the second largest value, and so forth. The query image is classified according to the weighted sum of similarities between its local features and all prototypes, with weights decaying exponentially to account for the decreasing influence of prototypes.
Zhou et al.~\cite{zhou2022hierarchical} proposed the Progressive Hierarchical Refinement (PHR) method to update prototypes iteratively using all novel data. In each iteration, support images and a random subset of query images are embedded into features at local, global and semantic levels, and a loss function defined over these hierarchical features is used to refine prototypes for better inter-class separability. As each update is based on a random subset of queries, the method is less likely to overfit to noisy query samples, though it implicitly assumes the availability of a large number of queries. 

Sun et al.~\cite{sun2021amortized} proposed to treat prototypes as random variables. The posterior distributions of latent class prototypes are learned by using amortized variational inference, a technique which enables prototype learning to be formulated as a probabilistic generative model without encountering severe computational and inferential difficulties.

\subsubsection{Distribution or subspace-based representations}
Considering that single point-based metric learning is sensitive to noise, Zhang et al. \cite{Zhang_2019_ICCV} proposed a variational Bayesian framework for few-shot learning and used the Kullback-Leibler divergence to measure the distance of samples. The framework can compute the confidence that a query image is assigned into each class by estimating the distribution of each class based on a neural network. 

Simon et al.~\cite{simon2020adaptive} proposed Deep Subspace Network~(DSN) to represent each class using a low-dimensional subspace, constructed from support samples via singular value decomposition. Query samples are classified according to the nearest subspace classifier, that is to assign the query to the class which has the shortest Euclidean distance between the query and its projection onto the class-specific subspace. The method is shown to be more robust to noises and outliers than Prototoypical Network. 

\subsection{Learning distance or similarity measures} \label{Sec:Metric learning}
Methods reviewed in Sections \ref{sec: feature embedding} and \ref{sec: prototype} focus on learning a discriminative feature embedding or obtaining an accurate class representation. For classification, they mostly adopt a fixed distance or similarity measure, such as the Euclidean distance \cite{snell2017prototypical} and the cosine similarity \cite{vinyals2016matching}. More recently, researchers propose to learn parameters in these fixed measures or define novel measures so as to further improve the classification accuracy. Moreover, considerable effort has been made to learn similarity scores by using fully-connected neural networks or Graph Neural Networks~(GNNs).

\subsubsection{Learning or selecting an analytical distance or similarity measure}

In TADAM~\cite{oreshkin2018tadam}, Oreshkin et al. mathematically analyzed the effect of metric scaling on the loss function. Since then, many works tune the scaling parameter via cross-validation~\cite{dong2020learning,liu2020negative}. Zhu et al.~\cite{zhu2021temperature} proposed to use two different scaling parameters for the ground-truth class and other classes to enforce the same-class distance is much smaller than the different-classes distance. Moreover, the scaling parameters are gradually tuned every few episodes, which implements the idea of self-paced learning to learn from easy to hard. Chen et al.~\cite{chen2020variational} proposed to learn the scaling parameter in a Bayesian framework. By assuming a univariate or multivariate Gaussian prior and applying the stochastic variational inference technique for approximating the posterior distribution, a scaling parameter or a scaling vector can be learned respectively, which scales the distance equally over all dimensions or differently for each dimension. Task-specific scaling vectors can also be learned by learning a neural network from the task to variational parameters. 

The traditional Mahalanobis distance decorrelates and scales features using the inverse of the covariance matrix. In Simple CNAPS~\cite{bateni2020improved}, after extracting features using the architecture of Conditional Neural Adaptive Processes (CNAPS)~\cite{requeima32fast}, the classification is performed based on the Mahalanobis distance between query instances and class prototypes. Task-specific class-specific covariance matrices are estimated as convex combinations of sample covariance matrices estimated from instances of the task and instances of the class, regularized toward an identity matrix. 
Transductive Episodic-wise Adaptive Metric (TEAM) \cite{Qiao_2019_ICCV} learned task-specific metric from support and query samples. TEAM contains three modules, namely a feature extractor, a task-specific metric module, and a similarity computation module. The task-specific metric module learns a Mahalanobis distance to shrink the distance between similar pairs and enlarge the distance between dissimilar pairs, following the objective function of the pioneering metric learning method~\cite{xing2002distance}. 

Nguyen et al.~\cite{kampffmeyersen} proposed a dissimilarity measure termed SEN, which combines the Euclidean distance and the difference in the $L_2$-norm. Minimizing this measure will encourage feature normalization and consequently benefit the classification performance~\cite{zheng2018ring}. DeepEMD \cite{zhang2020deepemd} combined a structural distance over dense image representations, Earth Mover's Distance (EMD) and convolutional feature embedding to conduct few-shot learning. The optimal matching flow parameters in EMD and the parameters in the feature embedding are trained in an end-to-end fashion. Xie et al.~\cite{xie2022joint} introduced the Brownian Distance Covariance~(BDC) metric, a new distance measure founded on the characteristic function of random vectors. The metric has a closed-form expression for discrete feature vectors and can be computed easily by first computing the BDC matrix for every image and then calculating the inner product between two BDC matrices. The computation of BDC matrices also only involves standard matrix operations and can be formulated as a pooling layer, thus endowing the method with high computational efficiency and ease of integrating with other few-shot classification methods.

\subsubsection{Learning similarity scores via neural networks} \label{sec:learn_CNN_metric}

The Relation Network~\cite{sung2018learning} is the first work of introducing a neural network to model the similarity of feature embeddings in few-shot learning. It consists of an embedding module and a relation module. The embedding module builds on convolutional blocks for mapping original images into an embedding space, and the relation module consists of two convolutional blocks and two fully-connected layers for computing the similarity between each pair of support and query images. The learnable similarity measure enhances the model flexibility. Li et al.~\cite{li2020bsnet} pointed out that a single similarity measure may not be sufficient to learn discriminative features for fine-grained image classification and thus proposed the Bi-Similarity Network~(BSNet) which complements the relation module by a cosine module, forcing features to adapt to two similarity measures of diverse characteristics and consequently generating a more compact feature space. Relation Network and subsequent methods all use class labels to form binary supervision, indicating whether the image pair comes from the same class. Zhang et al.~\cite{zhang2021rethinking} argued that such binary relations are not sufficient to capture the similarity nuance in the real-world setting and therefore proposed a new method termed Absolute-relative Learning~(ArL) which, in addition to binary relations, constructs continuous-valued relations from attributes of images, such as colors and textures.

Different from Relation Network, Semantic Alignment Metric Learning (SAML)~\cite{Hao_2019_ICCV} adopted the Multi-Layer Perceptron (MLP) network for computing the similarity score. Specifically, SAML contains a feature embedding module and a semantic alignment module. In the semantic alignment module, a relation matrix at the level of local features is first computed by using fixed similarity measures and an attention mechanism, and then fed into a MLP network which outputs the similarity score between the query and the support class. 

Recently, some researchers adopt Graph Neural Networks (GNNs) to implement few-shot classification. Like the above reviewed works, GNN-based methods also use a neural network to model the similarity measure, while its advantage lies in the rich relational structure on samples \cite{kim2019edge}. Garcia et al. \cite{garcia2018few} proposed the first GNN-based neural network for few-shot learning, short for GNN-FSL here. It contains two modules, a feature embedding module and a GNN module. In the GNN module, a node represents a sample, and more specifically, equals the concatenation of features of the sample and its label. For a query sample, its initial label in the first GNN layer uses uniform distribution over $K$-simplex ($K$ is number of classes in few-shot learning), and its predicted label in the last GNN layer is used for computing the loss function.  

Like GNN-FSL, Edge-labeling Graph Neural Network (EGNN)~\cite{kim2019edge} also contains a feature embedding module and a GNN module with three layers. However, rather than labeling nodes, EGNN learns to label edges in GNN layers so that it can cluster samples explicitly by employing the intra-cluster similarity and inter-cluster dissimilarity. In EGNN, each GNN layer has its own loss function that is computed based on edge values in the layer, and the total loss is the weighted sum of loss functions of all GNN layers. The Transductive Relation-Propagation graph neural Network (TRPN)~\cite{ma2020transductive} explicitly modeled the relation of support-query pairs by treating them as graph nodes. After relation propagation, a similarity function is learned to map the updated node to a similarity score, which represents the probability that the support and query samples are of the same class. The class with the highest sum of scores is the predicted class. The Hierarchical Graph Neural Network (HGNN)~\cite{chen2021hierarchical}, aimed at modeling the hierarchical structure within classes, first down-samples support nodes to build a hierarchy of graphs and then performs up-sampling to reconstruct all support nodes for prediction.

The previous GNN-based methods focus simply on the relation between a pair of samples. In Distribution Propagation Graph Network (DPGN)~\cite{yang2020dpgn}, the global relation between a sample and all support samples is considered by generating a distribution feature from the similarity vector. A dual complete graph is built to proceed sample-level and distribution-level features independently, and a cyclic update policy is used to propagate between the two graphs. Information from the distribution graph refines sample-level node features and hence improves the classification based on their edge similarities.

\section{Challenges and Future Directions}\label{trends}

Even though few-shot metric learning methods have achieved the promising performance, there remains several important challenges that need to be dealt with in the future.

\paragraph{1. Improving generalized feature learning on few samples}

Regarding feature learning, in the existing few-shot metric learning methods, or even in the entire few-shot learning methods, researchers mostly try to learn discriminative feature based on the attention mechanism, data augmentation, multi-task learning and so on. To learn feature with good generalization ability from few labeled examples, new ways of evaluation and feature learning need to be developed.

\paragraph{2. Rethinking the use of episodic training strategy} While episodic training is a common practice to train metric learning methods in the few-shot learning setting, it is rigid to require each training episode to have the same number of classes and images as the evaluation episode; in fact, \cite{snell2017prototypical} observed the benefit of training with a larger number of classes. Moreover, the model gets updated after receiving an episode without regard to its quality and thus is prone to poorly sampled images like outliers. \cite{fei2021melr} is the first attempt to alleviate this problem by exploiting the relationship between episodes; more solutions are needed to identify episodes that are high-quality and useful to the novel task. Furthermore, we notice that episodic training can result in models that underfit the base dataset. One possible reason is that, by using episodic training, methods adopt continual learning on plenty of tasks sampled from the base dataset and suffer from catastrophic forgetting~\cite{mccloskey1989catastrophic,gidaris2018dynamic}, i.e., the model learned from previous tasks is supplanted after learning on a new task. Therefore, how to avoid this problem and enhance the model fitting ability of metric learning methods on both base and novel datasets remains a challenge.

\paragraph{3. Enhancing stability to support samples and robustness to adversarial perturbations and distribution shifts}
Despite the continuous improvement in classification accuracy, few-shot classification methods are vulnerable in various scenarios, hindering their usage in safety-critical applications such as medical image analysis. Prior works show that existing methods are non-robust to input or label outliers~\cite{lu2020robust}, adversarial perturbations (i.e., small, visually imperceptible changes of data that fool the classifier to make incorrect predictions) added to support~\cite{oldewage2021attacking} or query images~\cite{goldblum2020adversarially}, and distribution shift between support and query datasets~\cite{bennequin2021bridging}. In~\cite{agarwal2021sensitivity}, it is demonstrated that even non-perturbed and in-distribution support images can significantly deteriorate the classification accuracy of several popular methods. Further exploration of vulnerability in existing approaches and design of robust and stable models will be very valuable.

\paragraph{4. Developing metric learning methods for cross-domain few-shot classification} 
While base and novel datasets may come from different domains in practice, currently only few works focus on cross-domain few-shot classification. More recently and severely, \cite{guo2020broader} reported that all meta-trained methods, including the reviewed work~\cite{tseng2020cross}, are outperformed by the simple transductive fine-tuning in the presence of a large domain shift, specifically, when training on natural images and evaluating beyond them, such as on agriculture and satellite images. The difficulty is that the base data and the novel data usually have different metric spaces. Therefore, how to alleviate domain shift between the training and evaluation phases needs to be explored in the future.

\section{Conclusions}\label{Conclusion}
This paper presents a review of recent few-shot deep metric learning methods. We provide the definitions and a general evaluation framework for few-shot learning, then categorize and review representative approaches, and finally summarize the main challenges. Based on these challenges, several new directions can be further explored in the future.

\bibliographystyle{ieeetr}
\bibliography{reference}

\end{document}